\documentclass[conference]{IEEEtran}
\usepackage[a4paper,
            left=0.509in,
            right=0.509in,
            top=0.75in,
            bottom=1.69in]{geometry}
            
\usepackage{graphicx}
\usepackage{textcomp}
\usepackage{array,ragged2e}
\def\BibTeX{{\rm B\kern-.05em{\sc i\kern-.025em b}\kern-.08em
    T\kern-.1667em\lower.7ex\hbox{E}\kern-.125emX}}

\usepackage{colortbl}
\usepackage{makecell}
\usepackage{pgfplots}
\pgfplotsset{compat=1.17}
\usepackage{amsmath}
\usepackage{multicol}
\usepackage{multirow}
\usepackage{tabto}
\usepackage{wrapfig}
\usepackage{amssymb}
\usepackage{amsmath}
\usepackage{float}
\usepackage{graphicx}
\usepackage[misc,geometry]{ifsym}
\usepackage{booktabs}%
\usepackage{textcomp}
\usepackage{array,ragged2e}
\usepackage{pgfplots}
\usepackage{url}
\usepackage{lineno,hyperref}
\hypersetup{
  linkcolor  = red!60!black,
  citecolor  = blue!90!white,
  urlcolor   = violet!60!black,
  colorlinks = true,
}

\usepackage{orcidlink}
\usepackage{comment}
\usepackage{hyperref}
\usepackage{svg}
\usepackage{xcolor}

\usepackage{threeparttable}

\usepackage{pgf-pie}  

\begin{document}
\title{Brain Tumor Segmentation in MRI Images with 3D U-Net and Contextual Transformer}
\thanks{Emails:20521783@gm.uit.edu.vn;20521654@gm.uit.edu.vn; hieubt@ueh.edu.vn;thienntb@uit.edu.vn;vuong.nm@ou.edu.vn\\
\textsuperscript{*}Correspondence: thienntb@uit.edu.vn}
\author{\IEEEauthorblockN{  Thien-Qua T.Nguyen\textsuperscript{1,2},
                            Hieu-Nghia Nguyen\textsuperscript{1,2},
                            Thanh-Hieu Bui\orcidlink{0000-0003-0443-7800}\textsuperscript{3},
                            Thien B. Nguyen-Tat\orcidlink{0000-0002-4809-7126}\textsuperscript{1,2,*},
                            Vuong M. Ngo\orcidlink{0000-0002-8793-0504}\textsuperscript{4}
                            }
\IEEEauthorblockA{\textit{\textsuperscript{1}University of Information Technology, Ho Chi Minh City, Vietnam} \\
\textit{\textsuperscript{2}Vietnam National University, Ho Chi Minh City, Vietnam}\\
\textit{\textsuperscript{3}College of Technology and Design, University of Economics Ho Chi Minh City, Ho Chi Minh City, Vietnam}\\
\textit{\textsuperscript{4}Ho Chi Minh City Open University, Ho Chi Minh City, Vietnam}\\
\textit{\textsuperscript{*}Correspondence: thienntb@uit.edu.vn}
}
}    

\maketitle

\begin{abstract}
This research presents an enhanced approach for precise segmentation of brain tumor masses in magnetic resonance imaging (MRI) using an advanced 3D-UNet model combined with a Context Transformer (CoT). By architectural expansion CoT, the proposed model extends its architecture to a 3D format, integrates it smoothly with the base model to utilize the complex contextual information found in MRI scans, emphasizing how elements rely on each other across an extended spatial range. The proposed model synchronizes tumor mass characteristics from CoT, mutually reinforcing feature extraction, facilitating the precise capture of detailed tumor mass structures, including location, size, and boundaries. Several experimental results present the outstanding segmentation performance of the proposed method in comparison to current state-of-the-art approaches, achieving Dicescore of 82.0\%, 81.5\%, 89.0\% for Enhancing Tumor, Tumor Core and Whole Tumor, respectively, on BraTS2019.
\end{abstract}

\IEEEpeerreviewmaketitle

\section{Introduction}
\label{Sec:Introduction}

Brain tumors are abnormal growths of cells in the brain, which can be either malignant or benign. These tumors can significantly impact the patient's quality of life and health, especially when they grow rapidly and spread to other areas of the brain and spinal cord. Imaging methods like X-rays and MRI are used to detect brain tumors, but not all of them can show the full details of the tumor \cite{chahal2020survey}. This increases the importance of using modern diagnostic methods, including artificial intelligence, to identify and classify brain tumors. Automating this procedure not only reduces costs and  saves time but also lightens the workload for staff and healthcare systems, promoting efficiency and resource conservation. With their profound impact on health and life, as well as the increasing number of cases, brain tumors are not just a medical issue but also an economic and social challenge.

\begin{figure}[!t]
    \centering
    \includegraphics[scale=0.40]{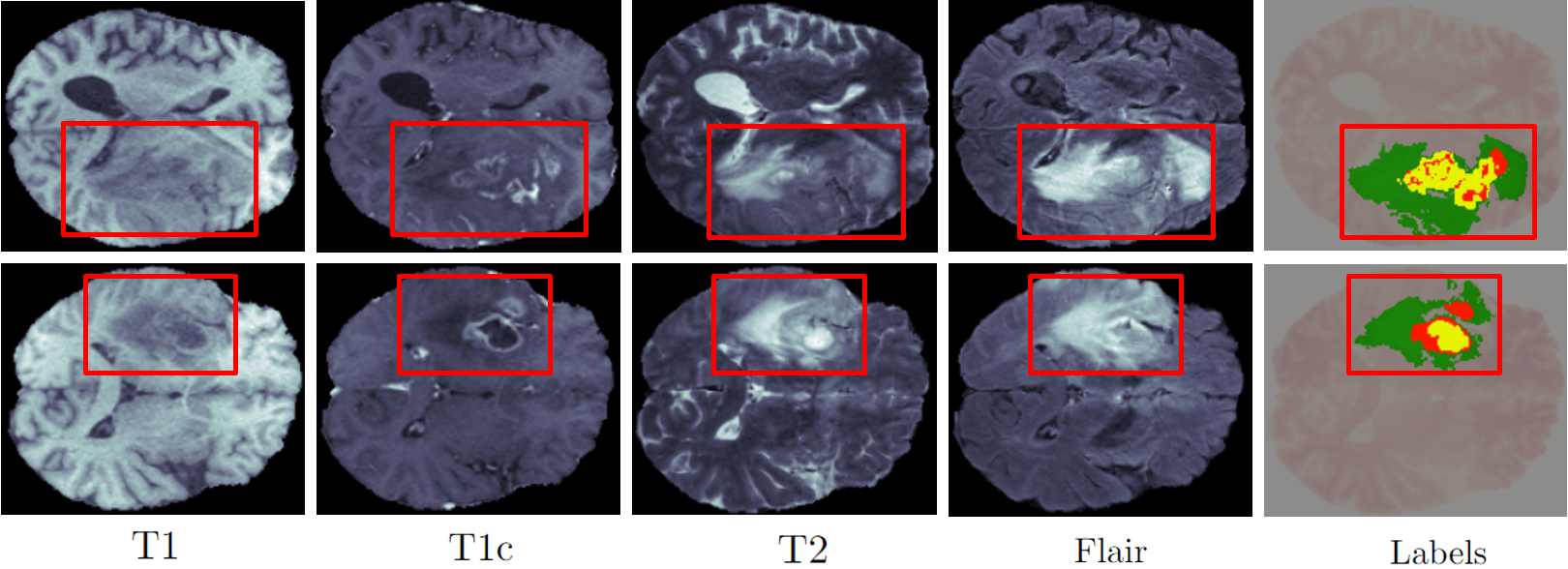}
    \caption{This figure displays modalities in two distinct cases, illustrating how these various modalities distinctly delineate different regions of tumor. The blue, red, yellow denote  tumor core, enhancing tumor and peritumoral edema, respectively}
    \label{fig1_samples}
\end{figure}

MRI, a widely used medical imaging technology, is commonly employed in clinical settings to assess brain tumors. Four main MRI modalities includeT1-weighted (T1), T2-weighted (T2), contrast-enhanced T1-weighted (T1c) and fluid attenuation inversion recovery (FLAIR) producing high-quality images of soft tissue abnormalities in the brain. The combination of these modalities enhances the accuracy of tumor segmentation, as depicted in Fig.\ref{fig1_samples}, where images from different modalities offer complementary information and mutual support.

The Transformer was first initially proposed by Vaswani et al. \cite{bib1}, an influential network architecture that represents a substantially advancement in deep learning and natural language processing. In the medical domain, the Transformer has opened up new opportunities in utilizing artificial intelligence in brain tumor segmentation from medical images.  By integrating attention mechanisms and learning  from large-scale data, the Transformer has become a strong tool for precisely and efficiently detecting and segmenting brain tumors \cite{bib2,bib3}. Transformer-based methods hold promise in addressing challenges in tumor segmentation, enhancing accuracy and reliability in the segmentation process.

Reference to the CT imaging study \cite{bib4}, we have taken inspiration and further expanded upon the research to provide more comprehensive and updated insight of brain tumor segmentation. Our study introduces a technique utilizing a 3D U-Net model, which has been enhanced and combine with a Transformer specifically for MRI images. By incorporating long-range information throughout the entire space, this advanced method allows for the precise identification and localization of tumor subregions. To achieve this, we have developed a Transformer-based model called Context Transformer \cite{bib5}, which incorporates an improved attention mechanism to explore features and contextual information. This innovative approach not only improve the accuracy of segmentation but also ensures efficiency and effectiveness in the process. This represents a notable progress in medical image segmentation, potentially enhancing diagnosing and treating patients.

The key contributions of this paper are as follows:
\begin{itemize}

\item The Contextual Transformer extended to 3D integrates with 3D UNet model to exploit rich contextual information in MRI images.

\item The proposed model has extended the architecture from the baseline, harmonizing tumor specific features sourced from CoT to extract important attributes. This comprehensive synthesis empowers accurate division of the complete tumor structure, including its location, size, shape, and boundaries. The best scoring results on the BraTS2019 dataset are 82.0\%, 81.5\%, 89.0\% respectively, for labels corresponding to Enhancing Tumor, Tumor Core, and Whole Tumor.

\end{itemize}
The rest of paper is structured as: Section 2 reviews related works. Section 3 show the proposed method. Section 4 delineates the experimental results, while Section 5 provides a summary of the paper's content and future work.

\section{Related work}
\label{sec:Related_work}
Image segmentation plays a crucial role in the healthcare field, particularly in diagnosing and treating diseases. Various techniques have been developed for segmenting brain tumor images \cite{bib6, bib7}, including both traditional machine learning (ML) methods and deep learning (DL) techniques. ML methods  such as Support Vector Machines \cite{bib8} and Graph Theory \cite{bib9}, have limitations in extracting statistical Information from large samples, resulting in weak segmentation performance. However, DL-based methods, particularly Convolutional Neural Network (CNN) based methods like 3D U-Net \cite{bib11} and Attention U-Net \cite{bib13}, have proven to be more effective in addressing this issue. These networks are capable of processing input images of any size and utilize decoding layers to adjust the size of feature maps to match the dimensions of the original image. CNN-based models with U-shaped architectures, have made significant advancements and demonstrated great potential in 2D and 3D image segmentation tasks. Nonetheless, the positioning of convolutional layers within the network architecture may lead to the ignore of long-range information correlations. Research \cite{xia2020md} has indicated that achieving good segmentation results requires a model that can simultaneously extract both local details and global semantic information interactions.

Transformer-based methods can address above issue. Liu et al. introduced the Swin Transformer, utilizing self-attention mechanisms based on windows to decrease parameters and computations, while employing a shifted window mechanism to realize global dependencies. Furthermore, Lin et al., introduced DS-TransUNet, a Transformer architecture similar to Unet for segmentation of medical images, achieving performance comparable to state-of-the-art CNN-based methods \cite{liu2021swin, lin2022ds}. However, the Transformer neglects local structures by dividing the image into patches represented as tokens.

Targeting the weaknesses of both CNN-based and Transformer-based networks, combining these structures can complement each other to exploit long-range spatial relationships. TransUnet\cite{bib22} marks the debut of Transformer in CNN. The CNN block of this work is implemented before Transformer. Then, features are restored by sampling through each layer. Achieving accurate image segmentation requires a significant amount of computational power and  overall data volume increase significantly when processing 3D data.


\section{Method}
\label{sec:Method}
In this paper, we introduce a network depicted in Fig. \ref{Figure2_model_architecture}, based on previously introduced transformer modules but incorporates enhanced channel attention modules. This allows us to explore spatial information and contextual in MRI images comprehensively, exploit features thoroughly, and improve the representation of various tumor regions. Consequently, we address the challenge of accurately capturing detailed information about both the entire tumor architecture and the characteristics of individual subregions, thereby enhancing segmentation accuracy. Our proposed network contains two main components: Fig. \ref{Figure2_model_architecture}a: the 3D-UNet backbone and Fig. \ref{Figure2_model_architecture}b: the 3D context-aware transformer module within  encoder-decoder

\subsection{3D Contextual Transformer (CoT)}

\begin{figure*}[!t]
    \centering
    \includegraphics[scale=0.267]{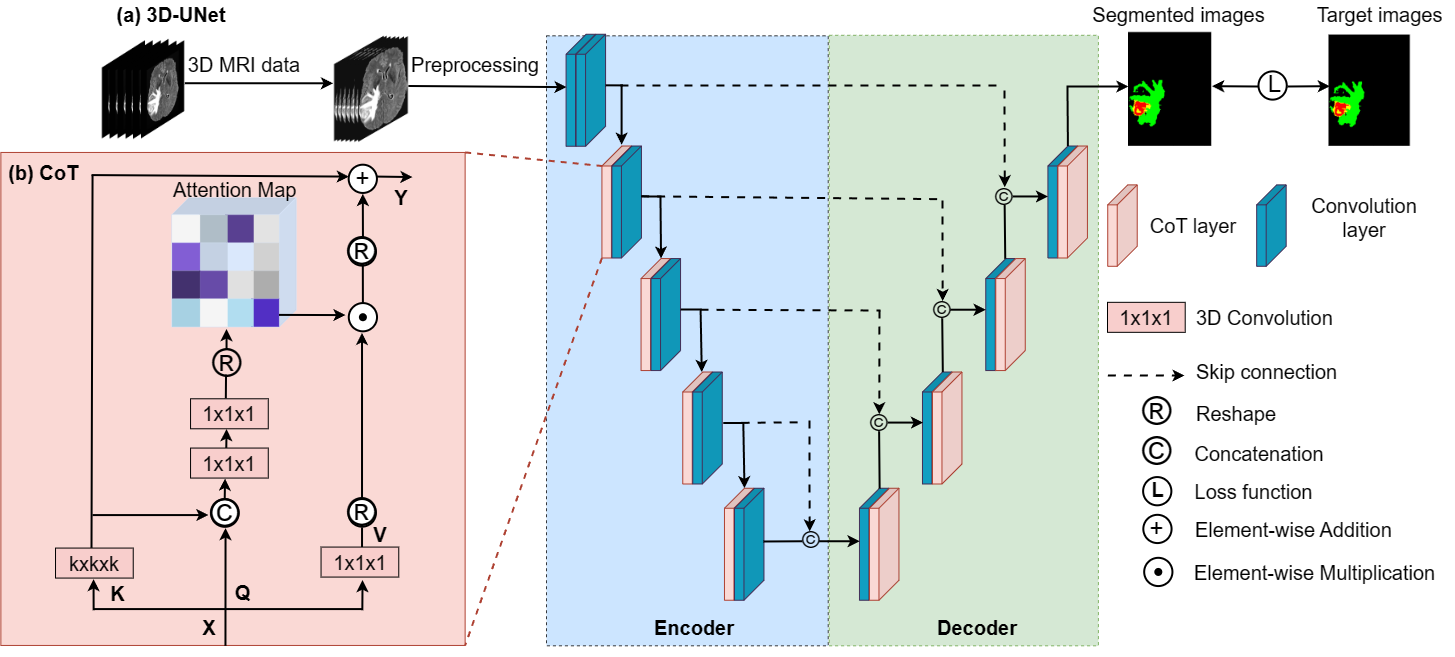}
    \caption{The architectural framework outlines our proposed approach for segmenting brain tumors from MRI images, utilizing the 3D-UNet architecture. \textbf{(a)}. Represents the 3D-UNet backbone model. \textbf{(b)}. Depicts the 3D contextual transformer (CoT) block directly linked to the convolutional layer}
    \label{Figure2_model_architecture}
\end{figure*}

The 2D contextual transformer module, aimed at utilizing contextual information within input features, was initially proposed by Li et al. \cite{bib5}, limited at 2D feature maps. In order to overcome this constraint, a 3D Contextual Transformer block is proposed, as depicted in Fig. \ref{Figure2_model_architecture}b. This CoT block integrates the utilization of contextual information and self-attention learning within a unified framework. It extensively leverages contextual information among adjacent keys to effectively support the self-attention learning process, thus improving the representation capability of the resulting output feature maps.


Initially, the 3D input feature map \textit{X $\in$ $\mathbb{R}^{H\times W\times D\times C}$}, with dimensions \textit{(H, W, D)} and \textit{C} channels, undergoes transformation into keys \textit{K}, values \textit{V} and queries \textit{Q} using learned embedding matrices \textit{$W_K$}, $W_V$ and $W_Q$, respectively. Subsequently, contextual information \textit{$K^1 \in \mathbb{R}^{H\times W\times D\times C}$} for the input \textit{X} is derived by applying a \textit{k × k × k} convolution across all adjacent keys to contextualize each key representation \textit{K}. This convolution inherently captures static contextual information among local neighboring keys. Next, the contextual keys \textit{$K^1$} and queries \textit{Q} are merged, and the resultant matrix undergoes two consecutive \textit{1 × 1 × 1} convolutions to produce the attention matrix \textit{A}.  The equation for this process is as follows.

\[
A = \left[ K^1, Q \right] W _\theta W_\delta  \tag{1}
\] where, $W_\theta W_\delta$ are learnt parameters.

In the subsequent step, dynamic contextual representations are obtained by performing element-wise multiplication between the feature map \textit{A} and the values \textit{V}

\[ K^2 = V * A \tag{2} \]

The CoT block produces the final output 
(Y) by merging the static context \textit{$K^1$} with the dynamic context \textit{$K^2$}.


\subsection{3D-UNet model and Loss function}

The 3D UNet model is a neural network variant commonly employed in medical image processing, especially for the segmentation of 3D medical scans like MRI or CT images. Derived from the U-Net architecture \cite{u-net}, a widely-used deep neural network in medical image analysis and segmentation, the 3D UNet model facilitates high-precision segmentation in 3D space. It achieves this by integrating down-sampling and up-sampling layers to analyze spatial information extracted from original images.

The Dice Loss has become increasingly popular as a loss function in semantic segmentation tasks. Its purpose is to measure and regulate the intersection between ground truth and predictions by optimizing the Dice coefficient directly. Within the module, both Dice Loss and cross-entropy loss are utilized to optimize the parameters. The definition of Dice Loss is as follows:

\[
\mathcal{L}_{dice}(y, \hat{y}) = 1 - \sum_{c\in\Omega_c} \frac{2 \cdot \sum_{i=1}^N y_i^c \hat{y}_i^c + \epsilon} {\sum_{i=1}^N (y_i^c)^2 + \sum_{i=1}^N (\hat{y}_i^c)^2 + \epsilon} \tag{3} 
\]and cross-entropy loss function is defined as follows:

\[
\mathcal{L}_{CE}(y, \hat{y}) =- \sum_{c\in\Omega_c} \sum_{i=1}^N   y_i^c \log \hat{y}_i^c \tag{4}
\] where $\Omega_c$ = \{BG(background),NCR/NET,ED,ET\}. $y_i^c$ and $\hat{y}_i^c$ denote the ground truth and probability prediction of voxel $i$ on class $c$, respectively. $N = H\times W\times D$, $\epsilon = 1\times 10^{-5}$. Consequently, due to equations (3) and (4),  the ultimate loss function is a weighted combination of the Dice Loss and cross-entropy loss, as indicated by the formula:

\[
\mathcal{L}_{Seg}(y, \hat{y}) = \alpha  \mathcal{L}_{dice}(y, \hat{y}) + (1-\alpha) \mathcal{L}_{CE}(y, \hat{y}) \tag{5}
\] where $\alpha$ is a hyperparameter that regulates the impact of Dice loss and cross-entropy loss.

\section{Dataset and Experiments}
\label{sec:Classification_Analysis}

\subsection{Evaluation Metrics}

The accuracy of segmentation in this research is assessed by employing the Dice score and  Hausdorff distance (95\%) metrics to evaluate enhancing tumor region (label 4), regions within the tumor core (label 1, 4) and entirety of tumor region (label 1, 2, 4).

The formula for calculating the Dice Score is given by:
\[ \mathcal DiceScore = \frac{2TP} {FN + FP + 2TP} \tag{6} \], 
where TP represents the number of true positives, FN represents the number of false negatives, and FP represents the number of false positives. To measure the dissimilarity between the actual surface of a region and the predicted region, the Hausdorff95 distance metric is employed. This metric is particularly sensitive to the boundaries of the segmented region and formally defined as follows:

\[ HD95(T, P) = \max \left\{ \sup_{t \in T} d(t, P), \ \sup_{p \in P} d(T, p) \right\} \tag{7}\]

The supremum operator, denoted as $sup$, is used in the context where $t$ and $p$ represent points on the surface $T$ of the ground-truth region and the surface $P$ of the predicted region. The function $d(t,p)$ calculates the distance between $t$ and $p$.

\subsection{Implementation details}

\textbf{Datasets:} The proposed method is evaluated using a dataset BraTS2019, which is provided by Brain Tumor Segmentation (BraTS) challenge. For training purposes, BraTS2019 consists of 335 patient cases. The validation set comprises MRI scans from 125 cases, with labels that are unknown. To train our model, we only utilize the labeled data, splitting it into 80/20 for training and testing. These datasets consist of co-registered, skull-stripped and resampled MRI images at a resolution of $1mm^3$. Each sample contains four MRI brain sequence modalities, namely Flair, T1, T1c, and T2. All modalities are aligned within the same space and have a volume size 240 × 240 × 155 voxels.

\textbf{Preprocessing:} Resampling is not unnecessary two datasets since all modalities have already been co-registered into a unified space. However, to ensure consistent pixel values across the entire training set, z-score normalization is necessary for the input data to non-zero values both the medical images and labels. The initial image dimensions are 240×240×155×1, while the merged image dimensions are 240×240×155×4. Afterwards, the images are cropped into fixed-size patches of 128×128×128×4 by removing any extraneous background voxels.

\textbf{Training Details:} To evaluate the efficacy of the proposed model, we conducted sequential training on various combinations of models, which included the baseline model and the baseline integrated with the CoT model. All models were trained from scratch, the proposed method was based on PyTorch and utilized the NVIDIA Tesla P100 GPU for training with a batch-size 1. The training process involved an initial learning-rate $3e^{-4}$, a cosine scheduler was applied 100 epochs with 3-fold cross-validation. The network was trained using softmax cross-entropy loss, and the model was regularized using L2 Norm with a weight decay rate $1e^{-5}$. During the inference stage, a sliding window was employed, utilizing a patch-size  128 × 128 × 128.

\section{Results and Discussion}
\label{sec:Discussion}

\subsection{Ablation study}

To evaluate the performance of the transformer block on two datasets, we conducted experiments using a combined model, and then compared them to the baseline, against the same evaluation set for each dataset.

Contextual Transformer (CoT): According to metrics from Table \ref{table1_results}, the combination of baseline + CoT demonstrates a considerable improvement in Dicescores for ET, achieving 82.0\% (an increase of 5.6\%) in comparison to the baseline. Besides that, the TC and WT label achieve 81.5\% and 89.0\%, respectively, resulting in a mean Dicescore increase of 2.2\%.  Furthermore, there is also an enhancement in the maen HD95, reduced by 1.1mm on BraTS2019 evaluation set. The incorporation of CoT blocks has resulted in a significant decrease in segmentation errors in all areas of the tumor, providing strong evidence of its effectiveness in improving the ability to differentiate between tumor subregions and enhancing overall segmentation performance. Additionally, the 3D UNet+CoT model prioritizes the interaction of contextual information to further improve segmentation accuracy. This experiment showcases the model's capacity to reconstruct tumors with greater precision by exchanging information across various spatial image domains, leading to a clearer understanding of tumor characteristics such as location, shape, and boundaries. As a result, the integration of multimodal features becomes more feasible for reliable segmentation tasks.

\begin{figure}[!t]
    \centering
    \includegraphics[scale=1.0]{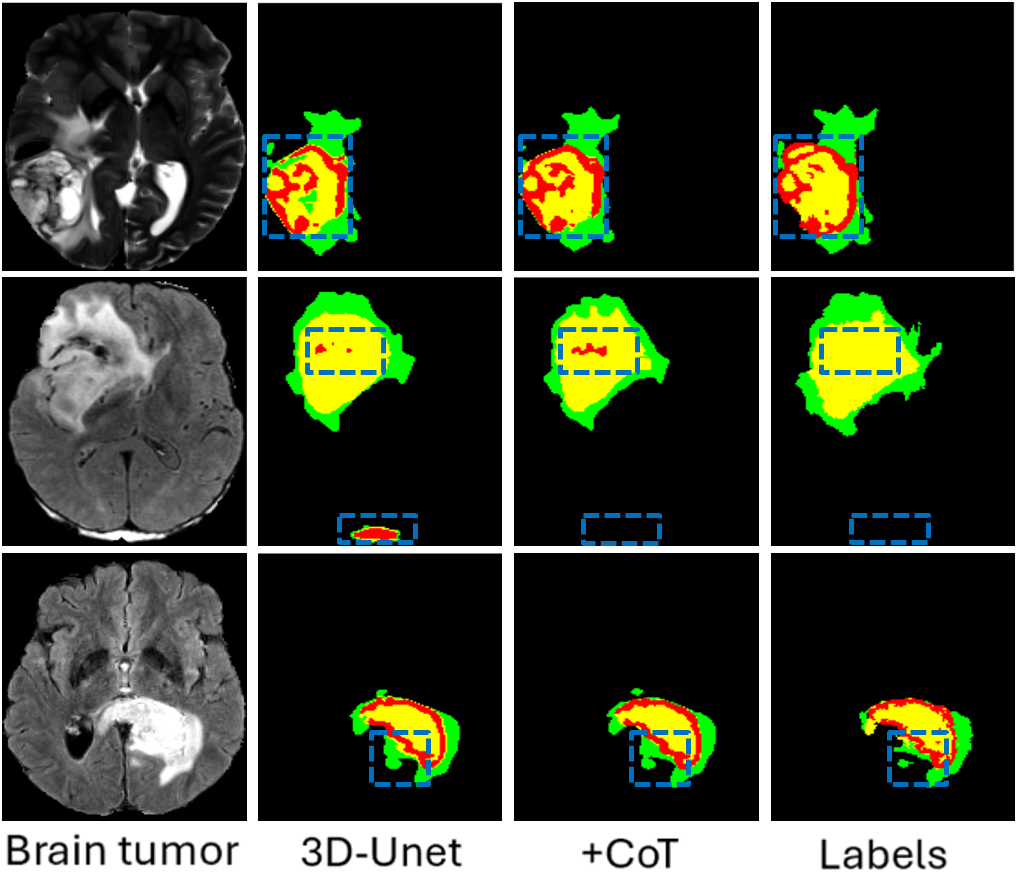}
    \caption{The differences between various components are visually compared, showcasing their effectiveness through good cases on the validation set BraTS2019. The variations are represented by dash-squares. The yellow, red, green regions denote the tumor core, the enhancing tumors and peritumoral edema, respectively}
    \label{fig3:ablatyStudy}
\end{figure}

Table \ref{table1_results} and Fig.\ref{fig_parameter} shows the parameter of the model combined with CoT significantly decreases, down to only 1.7M, indicating that the added transformer blocks have been used more consistently, helping to reduce memory and mitigate the risk of over-fitting. The proposed model architecture has been expanded to accommodate information synthesis needs, leading to an increase in training time. Furthermore, we show the segmentation results of various components in Fig.\ref{fig3:ablatyStudy}. Several case studies to illustrate the success of the segmentation according to the structures of individual tumors. These cases demonstrate insignificant differences between structures, as all are segmented very well.

\subsection{Evaluation of the influence  of each modality}

\begin{table*}[!t]
\centering
\caption{The performance of the models on the BraTS2019 validation set with 3-fold (mean\text{\footnotesize $\pm std$})}
\begin{tabular}{w{l}{1.5cm} w{c}{1.2cm} w{c}{1.2cm}w{c}{1.2cm}w{c}{1.2cm}w{c}{1.2cm}w{c}{1.2cm}w{c}{1.2cm}w{c}{1.2cm}}
\toprule
\multirow{2}{*}{Model} &  \multicolumn{4}{c}{Dice score (\%)} & \multicolumn{4}{c}{HD95 (mm)}\\\cmidrule(lr){2-5} \cmidrule(lr){6-9}
                        & ET   & TC    & WT   &  Avg. & ET & TC & WT & Avg. \\\midrule

3D-Unet   &  76.4\tiny $\pm 0.3 $ & 81.2\tiny $\pm0.5$  & 88.6\tiny$\pm 0.7$ & 82.0\tiny$\pm 0.5$ & 5.6\tiny$\pm 0.4$ & 7.6\tiny$\pm 0.5$ &  7.9\tiny$\pm 0.3$ & 7.0\tiny$\pm 0.2$ \\

+CoT   &  \textbf{82.0\tiny$\pm 0.5 $} & \textbf{81.5\tiny$\pm 0.6$}  & \textbf{89.0\tiny$\pm 0.8 $} & \textbf{84.2\tiny$\pm 0.6$} & \textbf{3.7\tiny$\pm 0.4$} & \textbf{7.4\tiny$\pm 0.4$} &  \textbf{6.7\tiny$\pm 0.5$} & \textbf{5.9\tiny$\pm 0.2$}\\

\hline
\end{tabular}
\label{table1_results}
\end{table*}

\begin{figure*}[!t]
\small
 \begin{center}
  \begin{tikzpicture}
   \pgfplotsset{height=36mm, width= 90mm, xlabel near ticks, ylabel near ticks, 
    plot 1/.style={blue!60!black,fill={rgb,255:red,0; green,100; blue,200},mark=none},%
    plot 2/.style={green!60!black,fill={rgb,255:red,130; green,255; blue,130},mark=none},%
    plot 3/.style={blue,fill=blue!30!white,mark=none},%
    plot 4/.style={white!50!black,fill=white!30!white,mark=none},%
    plot 5/.style={brown!60!black,fill=brown!30!white,mark=none},%
    plot 6/.style={red,fill=red!30!white,mark=none},%
   }
    
   \begin{axis}[
    name=left axis,
    ybar,
    bar width=4.2pt, 
    enlargelimits=0.19,
    legend style={/tikz/every even column/.append style={column sep=0.2cm},
    draw=none, fill=none, font=\footnotesize, legend columns=-1, anchor = south, xshift= 3mm, yshift= -33mm}, 
    y label style={align=center},
    ylabel={Dice score (\%)},
    every axis y label/.style={
    	at={(ticklabel* cs:1.05)},
    	anchor=south, xshift= 5mm,
    },
    ymin=0.7, ymax=0.9,
    symbolic x coords={ET, TC, WT, Avg.},
    xtick=data,
    every node near coord/.append style={font=\scriptsize},  
    ]

    \addplot[plot 1] coordinates{(ET, 0.732) (TC,0.739) (WT, 0.876) (Avg., 0.782)};
    \addplot[plot 2] coordinates{(ET,0.739) (TC,0.766) (WT, 0.887) (Avg., 0.797)};
    \addplot[plot 3] coordinates{(ET,0.760) (TC,0.772) (WT, 0.888) (Avg., 0.807)};
    \addplot[plot 4] coordinates{(ET,0.782) (TC,0.789) (WT, 0.895) (Avg., 0.822)};
    \addplot[plot 5] coordinates{(ET,0.788) (TC,0.823) (WT, 0.893) (Avg., 0.835)};
    \addplot[plot 6] coordinates{(ET,0.820) (TC,0.815) (WT, 0.890) (Avg., 0.842)};

	\legend{Kiu-Net \cite{bib23}, V-Net \cite{bib20}, Attention U-net \cite{bib13}, TransUNet \cite{bib22}, CGA U-Net \cite{bib24}, Ours}
   \end{axis}

   \begin{axis}[
    name=center axis,
    at=(left axis.east),
    xshift= 1.5cm,
	yshift= -1cm,
    ybar,
    bar width=4.2pt,
    enlargelimits=0.19,
    legend style={/tikz/every even column/.append style={column sep=0.2cm},
    draw=none, fill=none, font=\footnotesize, legend columns=-1, anchor = south, xshift= 3mm, yshift= -33mm}, 
    y label style={align=center},
    ylabel={HD95 (mm)},
    every axis y label/.style={
    	at={(ticklabel* cs:1)},
    	anchor=south, xshift= 5mm,
    },
    ymin=3.0, ymax=10.0,
    symbolic x coords={ET, TC, WT, Avg.},
    xtick=data,
    every node near coord/.append style={font=\scriptsize},  
    ]

    \addplot[plot 1] coordinates{(ET, 6.3) (TC,9.9) (WT, 8.9) (Avg., 8.4)};
    \addplot[plot 2] coordinates{(ET,6.1) (TC,8.7) (WT, 6.3) (Avg., 7.0)};
    \addplot[plot 3] coordinates{(ET,5.2) (TC,8.3) (WT, 7.8) (Avg., 7.1)};
    \addplot[plot 4] coordinates{(ET,4.8) (TC,7.4) (WT, 6.7) (Avg., 6.3)};
    \addplot[plot 5] coordinates{(ET,3.3) (TC,6.7) (WT, 5.8) (Avg., 5.3)};
    \addplot[plot 6] coordinates{(ET,3.7) (TC,7.4) (WT, 6.7) (Avg., 5.9)};
   \end{axis}
   
  \end{tikzpicture}

 \end{center}
 \vspace{-2mm}
 \caption{Performances comparison with some SOTA on the validation set BraTS2019. All metrics are provided by the author}
 \label{fig:compare_SOTA}
\end{figure*}
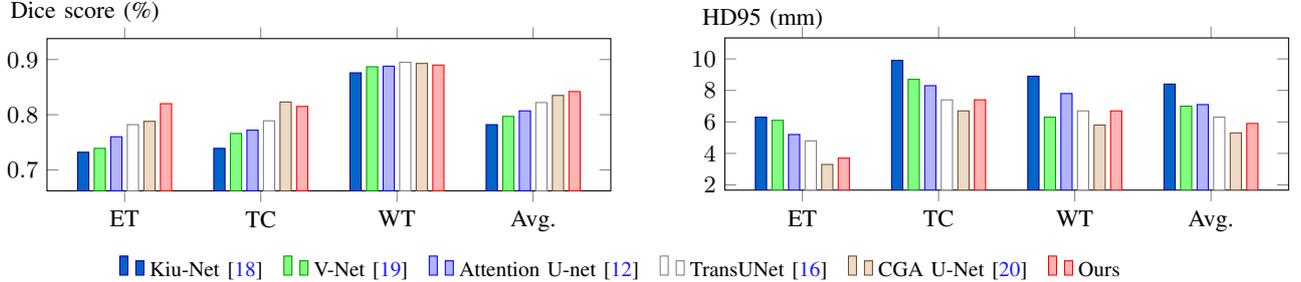


\begin{figure}[!t]
\small
 \hspace{1cm}
  \begin{tikzpicture}
   \pgfplotsset{height=36mm, width= 55mm, xlabel near ticks, ylabel near ticks, 
    plot 1/.style={blue!60!black,fill={rgb,255:red,255; green,215; blue,0},mark=none},%
    plot 2/.style={green!60!black,fill={rgb,255:red,230; green,240; blue,110},mark=none},%
   }
   \begin{axis}[
    ybar,
    bar width=6pt,
    enlargelimits=0.55,
    enlarge y limits={value=0.15,upper},
    legend style={at={(1.1, 2)}, /tikz/every even column/.append style={column sep=0.2cm}, draw=none, fill=none, font=\footnotesize, legend columns=1, anchor = south, xshift= 10mm, yshift= -33mm}, 
    y label style={align=center},
    every axis y label/.style={
    	at={(ticklabel* cs:1.05)},
    	anchor=south, xshift= 5mm,
    },
    ymin=0.0, ymax=11.5,
    symbolic x coords={3D UNet, +CoT},
    xtick=data,
    nodes near coords,
    nodes near coords align={vertical},
    every node near coord/.append style={font=\scriptsize}, 
    ]

    \addplot[plot 1 ] coordinates{(3D UNet, 2.51) (+CoT, 1.70) };
    \addplot[plot 2] coordinates{(3D UNet,6.83) (+CoT, 10.6)};

	\legend{Parameter (M), Training time (h)}
 
   \end{axis}
  \end{tikzpicture}
 \vspace{-2mm}
 \caption{ Comparison of parameter count and training time for each model (training time per epoch)}
 \label{fig_parameter}
\end{figure}
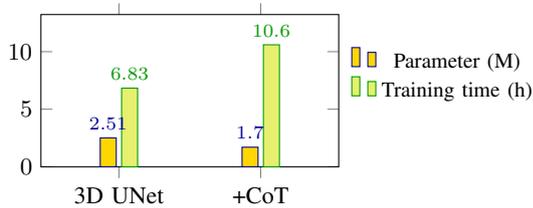

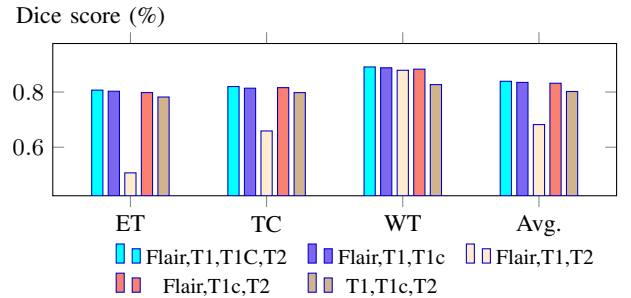
\begin{figure}[!t]
\small
 \begin{center}
  \begin{tikzpicture}
   \pgfplotsset{height=36mm, width= 90mm, xlabel near ticks, ylabel near ticks, 
    plot 1/.style={blue!80!black, fill={rgb,255:red,0; green,255; blue,255}, mark=none},
    plot 2/.style={blue!80!black, fill={rgb,255:red,123; green,104; blue,238}, mark=none},
    plot 3/.style={blue!80!black, fill={rgb,255:red,255; green,235; blue,205}, mark=none},
    plot 4/.style={ blue!80!black, fill={rgb,255:red,250; green,128; blue,114}, mark=none},
    plot 5/.style={blue!80!black, fill={rgb,255:red,210; green,180; blue,140}, mark=none},
   }
    
   \begin{axis}[
    ybar,
    bar width=4.2pt, 
    enlargelimits=0.19,
    legend style={at={(0.5,0.9)}, /tikz/every even column/.append style={column sep=0.2cm}, draw=none, fill=none, font=\footnotesize, legend columns=3, anchor = south, xshift= 3mm, yshift= -33mm}, 
    y label style={align=center},
    ylabel={Dice score (\%)},
    every axis y label/.style={
    	at={(ticklabel* cs:1.05)},
    	anchor=south, xshift= 5mm,
    },
    ymin=0.5, ymax=0.9,
    symbolic x coords={ET, TC, WT, Avg.},
    xtick=data,
    every node near coord/.append style={font=\scriptsize},  
    ]

    \addplot[plot 1] coordinates{(ET, 0.807) (TC,0.820) (WT, 0.891) (Avg., 0.839)};
    \addplot[plot 2] coordinates{(ET,0.803) (TC,0.814) (WT, 0.888) (Avg., 0.835)};
    \addplot[plot 3] coordinates{(ET,0.507) (TC,0.659) (WT, 0.879) (Avg., 0.682)};
    \addplot[plot 4] coordinates{(ET,0.798) (TC,0.816) (WT, 0.883) (Avg., 0.832)};
    \addplot[plot 5] coordinates{(ET,0.782) (TC,0.798) (WT, 0.827) (Avg., 0.802)};

	\legend{\text{Flair,T1,T1C,T2}, \text{Flair,T1,T1c}, \text{Flair,T1,T2}, \text{Flair,T1c,T2}, \text{T1,T1c,T2}}
 
   \end{axis}
  \end{tikzpicture}
 \end{center}
 \vspace{-5mm}
 \caption{Comparison of segmentation model performance, trained using different modalities, on the BraTS2019 evaluation set with the proposed model}
 \label{fig:resulsmodalities}
\end{figure}

In order to evaluate how different modalities affect the model's performance in segmenting tumors, we conducted sequential training of the proposed model (3D Unet+CoT) on the BraTS2019 evaluation set, excluding a modality at a time. The outcomes of this experiment are displayed in Fig.\ref{fig:resulsmodalities}, revealing that the omission of T1c has a significant negative impact on the TC and ET label, while the exclusion of Flair leads to a decrease in performance for the WT label. Clearly, each modality possesses its own unique characteristics. T1c plays a crucial role in enhancing the structural tumor's features, resulting in clearer and more distinguishable boundaries \cite{bib19}. The information conveyed by these features is instrumental in detecting, classifying core and enhancing areas of the tumor. Consequently, if T1c is not included, the model struggles to accurately discern the boundary features. The differentiation between cerebrospinal fluid and edema is aided by the suppression of water molecules in the FLAIR modality \cite{bib19}. Consequently, the FLAIR sequence has a significant impact on segmenting both the entire tumor region and overall tumor volume. T1 is valuable for differentiating normal tissues, however, it weakens the tumor's characteristics, while T2 is primarily utilized to differentiate edema regions and improve the signal in that specific area, providing valuable information for training the model. Each modality plays a crucial role and offers distinct features, resulting in optimal segmentation performance when combined.

\subsection{Comparison with state-of-the-arts}

To validate the efficacy of our proposed approach, we benchmark it against state-of-the-art (SOTA) segmentation approaches on the BraTS2019 dataset. The results are displayed in Fig. \ref{fig:compare_SOTA}. Our proposed model surpasses most current SOTA methods, especially excelling in dicescore for the ET label, achieving 82.0\%, with the average dicescore of 84.2\%. Nevertheless, although our approach performs well in the HD95, the CGA U-Net method \cite{bib24} has a slightly better. These results evidence of effectiveness, superiority and potentiality of our method over previous SOTA and recent Transformer-based methods (Attention U-net \cite{bib13}, TransUNet \cite{bib22}) on the validation set of BraTS2019.

\subsection{Error analysis}

\begin{figure}[!t]
    \centering
    \includegraphics[scale=1.15]{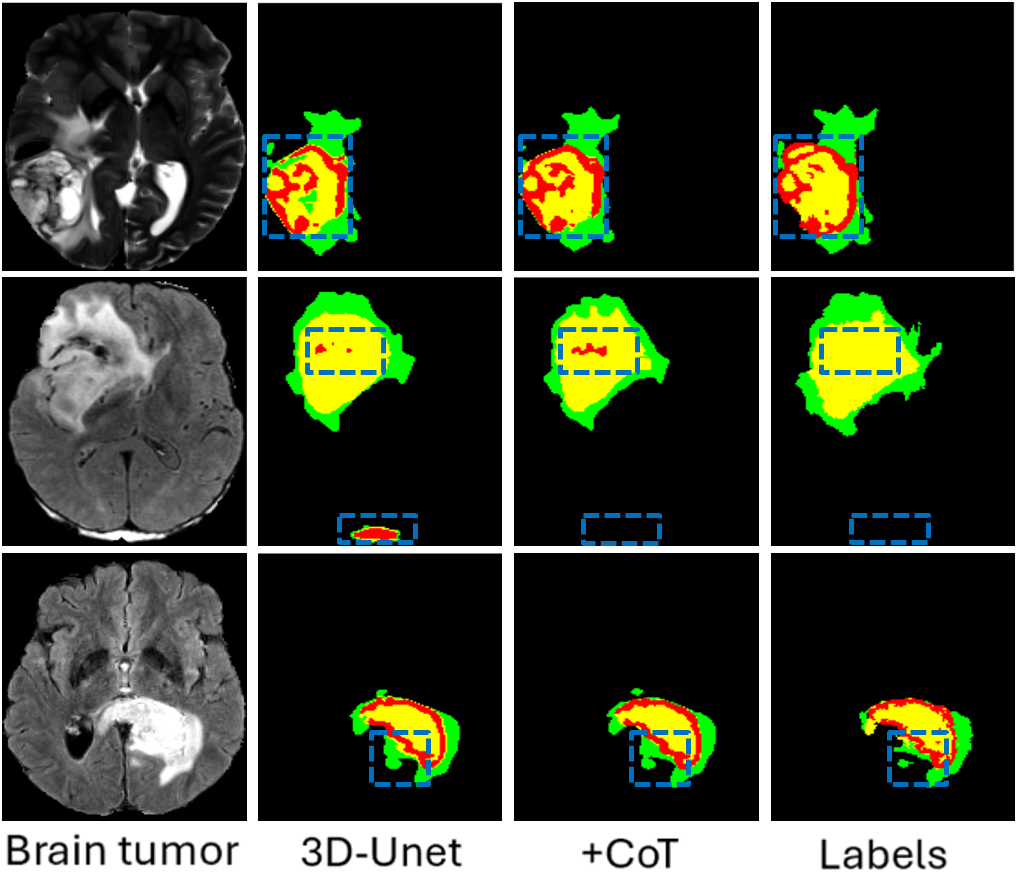}
     
    \caption{The differences between various components are visually compared, showcasing their effectiveness through bad samples on the validation set BraTS2019. The variations are represented by dash-squares. The yellow, red, green regions denote the tumor core, the enhancing tumors and peritumoral edema, respectively}
    \label{fig4:badcase}
\end{figure}

Although the approach performs well overall, it operates less efficiently in specific cases. For instance, in Fig.\ref{fig4:badcase}, the model outcomes segmentation does not entirely match the ground truth, but in comparison to the baseline, the 3D-Unet + CoT model provides relatively exact segmentation. In the first sample, several tumor cores and enhancing tumors remain not entirely accurate. On the second, the baseline model missegmented the enhancing tumor and was confused by a bright artifact below, which is a common noise scenario. In contrast, our model missegments only the enhancing tumor without being affected by the interfering noise. And, in the third, both models slightly misidentified the edges of the tumor that needs segmentation. Hence, the model can not precisely segment the tumor of boundaries, therefore missing essential tumor characteristics. The ability to detect and identify some small regions within complex tumors of the 3D UNet model is not truly accurate. This results in the loss of information concerning the tumor's boundaries with surrounding structures, leading to diagnostic errors. In comparison to the results of the baseline, the 3D-Unet+CoT model marginally enhances specific errors associated with size, shape, and location. The overall image of the tumor appears more comprehensive with less critical information loss. This significantly benefits providing reliable information to healthcare, thereby contributing to the formulation of optimal treatment decisions.

\section{Conclusion and future work}

In this paper, we introduce a robust technique for multimodal brain tumor segmentation from MRI images through integration with CoT to extend the baseline architecture, to improve segmentation accuracy. Specifically, CoT leverages tumor characteristics and contextual information by focusing on self-attention blocks, thereby enhancing the representation and synthesis of output information. As a CNN-Transformer architecture, it inherits the advantages of 3D-CNN in modeling local context and demonstrates the superior capability of Transformers in modeling long-range dependencies. Therefore, the 3D UNet+CoT model effectively synchronizes characteristics, supports each other in synthesizing crucial features. Consequently, this model can understand the complete tumor structure in detail and accuracy, including boundaries, locations, shapes, and sizes. Experimental results have validated the efficacy of the proposed approach, achieving Dicescores of 82.0\%, 81.2\%, and 88.6\% for the ET, TC, WT label on BraTS2019, outperforming several other state-of-the-art methods.

In the future, specialized medical pre-processing techniques could be implemented on MRI images to enhance segmentation performance. Additionally, using the 3D UNet model as a baseline requires considerable computational resources to process large datasets. Thus, optimizing computation becomes a research focus. Furthermore, this approach can also be utilized for medical image segmentation tasks associated with liver conditions such as fibrosis, hepatitis, or lung lesions. This creates opportunities to broaden the potential applications of study methodologies in the future within the domain of medical imaging.

\section*{Acknowledgment}
This research was supported by The VNUHCM-University of Information Technology's Scientific Research Support Fund.



\end{document}